\documentclass[letterpaper, 10 pt, conference]{ieeeconf}  

\IEEEoverridecommandlockouts                              

\usepackage[noadjust]{cite}
\usepackage{graphics} 
\usepackage{graphicx}
\usepackage{amsmath} 
\usepackage{amssymb}  
\usepackage{acronym}
\usepackage[font=footnotesize]{caption}
\usepackage{caption,subcaption,tabularx,booktabs}
\usepackage{array}
\newcolumntype{?}{!{\vrule width 0.2mm}}
\newcolumntype{A}{>{\centering}p{0.6cm}}
\usepackage{tikz} 
\pgfdeclareverticalshading{rainbow}{100bp}
{color(0bp)=(magenta);
 color(25bp)=(magenta);
 color(35bp)=(blue);
 color(45bp)=(cyan);
 color(55bp)=(green);
 color(65bp)=(yellow);
 color(75bp)=(red);
 color(100bp)=(red)}

 \usepackage{balance}

\newcommand{\bbm}{\begin{bmatrix}}
\newcommand{\ebm}{\end{bmatrix}}
\newcommand{\mbf}{\mathbf}

\newcommand{\mbs}[1]{{\boldsymbol{#1}}}

\newcommand{\beq}{\begin{equation}}
\newcommand{\eeq}{\end{equation}}
\newcommand{\bdis}{\begin{displaymath}}
\newcommand{\edis}{\end{displaymath}}
\newcommand{\bfone}{\mbf{1}}
\newcommand{\beqn}[1]{\begin{subequations}\label{eq:#1}\begin{eqnarray}}
\newcommand{\eeqn}{\end{eqnarray}\end{subequations}}
\newcommand{\x}{\mbf{x}}
\newcommand{\y}{\mbf{y}}
\newcommand{\Ss}{\mbf{S}}
\newcommand{\Ww}{\mbf{W}}
\newcommand{\fe}{\mbs{\psi}}
\newcommand{\ke}{h}
\newcommand{\T}{\mbf{T}}

\newcommand{\Sbig}{
\bbm
\ddots &             &             &           &        \\
\cdots & \bfone      &             &           &        \\
\cdots & \Ss_{K-2,1} & \bfone      &           &        \\
\cdots & \Ss_{K-1,2} & \Ss_{K-1,1} & \bfone    &        \\
\cdots & \Ss_{K,3}   & \Ss_{K,2}   & \Ss_{K,1} & \quad \bfone
\ebm
}

\acrodef{MAP}{Maximum A Posteriori}
\acrodef{SVD}{Singular Value Decomposition}
\acrodef{NEES}{Normalized Estimation Error Squared}
\acrodef{RMSE}{Root Mean Squared Error}
\acrodef{EM}{Expectation-Maximization}
\acrodef{ICP}{Iterative Closest Point}
\acrodef{GN}{Gauss-Newton}
\acrodef{IMU}{Inertial Measurement Unit}
\acrodef{EKF}{Extended Kalman Filter}

\title{\LARGE \bf
Towards Consistent Batch State Estimation Using a Time-Correlated Measurement Noise Model
}

\author{David J. Yoon and Timothy D. Barfoot \thanks{All authors are with the University of Toronto Institute
for Aerospace Studies (UTIAS), 4925 Dufferin St, Ontario, Canada.  \texttt{david.yoon@robotics.utias.utoronto.ca, tim.barfoot@utoronto.ca}}
}

\begin{document}

\maketitle
\thispagestyle{empty}
\pagestyle{empty}

\begin{abstract}
In this paper, we present an algorithm for learning time-correlated measurement covariances for application in batch state estimation. We parameterize the inverse measurement covariance matrix to be block-banded, which conveniently factorizes and results in a computationally efficient approach for correlating measurements across the entire trajectory. We train our covariance model through supervised learning using the groundtruth trajectory. In applications where the measurements are time-correlated, we demonstrate improved performance in both the mean posterior estimate and the covariance (i.e., improved estimator consistency). We use an experimental dataset collected using a mobile robot equipped with a laser rangefinder to demonstrate the improvement in performance. We also verify estimator consistency in a controlled simulation using a statistical test over several trials.
\end{abstract}

\section{Introduction}

In the research field of probabilistic robotics, we formulate state estimation using probability theory in order to handle the uncertainty of our sensor measurements \cite{thrun2005probabilistic}. Modern algorithms are capable of solving for the trajectory using all measurements in a batch solution \cite{Barfoot2017,Kaess-2017-106169}, resulting in accurate estimates in real-world applications.

\begin{figure}[!ht]
  \begin{minipage}[]{0.24\textwidth}
    \centering
    \text{covariance} \par
      \includegraphics[trim={3.55in 4.85in 3.48in 4.76in},clip]{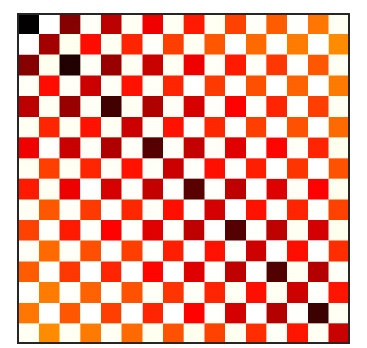}
  \end{minipage}%
  \hfill
  \begin{minipage}[]{0.24\textwidth}
    \centering
    \text{inverse covariance}\par
      \includegraphics[trim={3.55in 4.85in 3.48in 4.76in},clip]{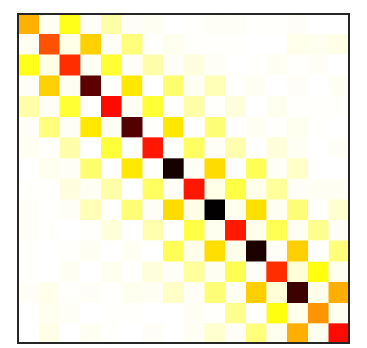}
  \end{minipage}

  \caption{A visualization of the sample measurement covariance (left) and its inverse (right) for a 2D landmark over 8 timesteps, showing the presence of time correlation in a real dataset. Darker colours indicate larger magnitudes and lighter colours are magnitudes closer to zero. The covariance appears dense, but its inverse appears banded, which our proposed method exploits.}
  \label{fig:lm_sample_cov}
  \vspace*{-0.2in}
\end{figure}

Estimator performance is affected by the uncertainty we assign to the noisy measurements, i.e., our sensor noise models. Ideally, we wish to learn our noise models from data. In most applications, where the noise is assumed Gaussian, common practice is to assume that measurements acquired at different times are corrupted by statistically independent noise. Using the groundtruth trajectory, the sample measurement covariance at the marginal level (i.e., at a single timestep) can then be applied as the noise model. However, the uncorrelated assumption is not always accurate (see Figure~\ref{fig:lm_sample_cov}). Not accounting for the correlations degrades the quality of the estimator and can lead to overconfident estimates \cite{merfels2017sensor}. Computing the sample covariance over the entire trajectory is also not tractable, as we often work with trajectories that are many thousands of timesteps in length.

In order to improve the state of the art in batch state estimation, we developed a data-driven approach for learning time-correlated noise models in a supervised learning fashion (i.e., using the groundtruth trajectory). By parameterizing our noise model as the inverse covariance, our method is computationally efficient for large trajectories. To the best of our knowledge, learning time-correlated measurement covariances for batch state estimation has not been demonstrated before in the mobile robotics research field. We verify that our proposed method results in a consistent batch estimator in a controlled simulation via a statistical test over several trials. We demonstrate improved performance in both the posterior mean and covariance (as gauged by estimator consistency) using an experimental dataset with time-correlated measurements.





\section{Related Work}
In applications where the sensor measurements are not time-correlated, several options exist for learning measurement covariances. The default method is to assume the noise characteristics remain constant throughout the trajectory, in which case the maximum likelihood sample covariance is a reasonable choice \cite{Barfoot2017}. If the groundtruth trajectory is unknown (i.e., unsupervised training), \ac{EM} can be applied to iteratively refine the trajectory estimate and measurement covariance \cite{shumway1982approach,ghahramani96,barfoot_ijrr20}.

In some applications, the noise characteristics may vary throughout the trajectory. An early approach for handling this problem was the adaptive Kalman Filter \cite{mehra1970identification,stengel1994optimal}, a reactive method that refines the sample covariance over a trailing window of frames. Rather than reacting, which may be delayed, a predictive approach, CELLO, was presented by Vega-Brown et al. \cite{vega2013cello}. By introducing known features that describe the measurements, local kernel estimates of the covariance were learned by weighting measurements based on feature similarity. Landry et al. \cite{landry2019cello} adapt CELLO to model the uncertainty of pointcloud registration for \ac{ICP}. Later work extended CELLO to unsupervised training using \ac{EM} \cite{vega2013celloem,peretroukhin2016probe}.

Deep learning presents another option for learning a varying noise model. Brossard et al. \cite{brossard2020ai} regress the noise variances of a vehicle model that penalizes lateral and vertical motion using a trailing window of \ac{IMU} data as input. Predicting the covariance matrices for richer sensor data has been demonstrated for cameras \cite{liu2018deep,russell2021multivariate} and for \ac{ICP} using lidars \cite{torroba2020pointnetkl,de2022deep}. Since processing rich sensor inputs is a strength of deep neural networks, a combined application with our proposed parameterization for correlated covariances is of interest for future work.


When the measurements are time-correlated, we may still assume an uncorrelated noise model and inflate covariances to mitigate estimator overconfidence. Inflation can be done manually, but may not be practical in complex real-world applications. An alternative is to apply the concept of \textit{covariance intersection} \cite{julier1997non}, which aims to fuse measurements with unknown correlation to result in a conservative (underconfident) estimate. A generalization of covariance intersection was applied to batch estimation (i.e., factor graphs) \cite{julier2012fusion,noack2015treatment}. As this approach tends to be more conservative than desired, the \textit{inverse covariance intersection} method was developed as a less-conservative alternative \cite{noack2019nonlinear}. While assuming an uncorrelated noise model has a computational benefit, our goal is to learn a noise model, including the correlations, in an automated way that results in a consistent estimator (i.e., neither underconfident or overconfident).

Methods for modelling time correlations in measurements exist for filter estimators. One approach \textit{augments} the state with the time-correlated noise. This has numerical issues due to removing the noise variable from the measurement model \cite{barshalom,bryson1968estimation}, for which Wang et al. \cite{wang2012practical} present a workaround. More recently, Russell and Reale \cite{russell2021multivariate} combined this method with a deep network. Their network outputs a mean and covariance estimate for visual odometry that was treated as a pseudomeasurement in an \ac{EKF}. An alternative to augmenting the state involves creating a \textit{differenced measurement} to cancel out the correlated portion of the noise \cite{barshalom,bryson1968estimation}. Variations of this method were demonstrated in simulation \cite{wang2012practical,chang2014kalman} and real data \cite{petovello2009,xu2020ins}. Wang et al. \cite{wang2014nonlinear} apply the idea to smoothing problems in simulation. Lee and Johnson \cite{lee2017state} propose to learn the correlations using Gaussian Process regression, which they demonstrate in simulation with the \ac{EKF}.

For batch estimation, similar to augmenting filters, Julier et al. \cite{julier2012multi} model and estimate the correlations as part of the state. They propose approximations to reduce the number of additional state variables and show results in simulation. This approach requires augmenting the state for each time-correlated sensor and a corresponding prior for those variables. We instead model a time-correlated measurement covariance over the entire trajectory efficiently by parameterizing a block-banded inverse covariance. Instead of requiring a priori knowledge of the correlations, we propose learning the noise parameters from data.






\section{Methodology}
\subsection{Problem Formulation}
In batch state estimation, our goal is to compute the posterior, $p(\mbf{x}|\mbf{y})$, where $\mbf{x} = \mbf{x}_{1:K} = \{\mbf{x}_1, \mbf{x}_2, \dots, \mbf{x}_K\}$ are our states at discrete times and $\mbf{y} = \mbf{y}_{1:K} = \{\mbf{y}_1, \mbf{y}_2, \dots, \mbf{y}_K\}$ are the noisy measurements. Note that $\x_k \in \mathbb{R}^D$ and $\y_k \in \mathbb{R}^M$. Using Bayes' rule, $p(\mbf{x}|\mbf{y}) = p(\mbf{y}|\mbf{x}) p(\mbf{x})/p(\mbf{y})$.
Taking the negative logarithm and dropping constants, the standard \ac{MAP} objective \cite{Barfoot2017} is
\begin{equation}
J = \underbrace{
  \frac{1}{2}\mbf{e}_y^T \mbf{R}^{-1} \mbf{e}_y - \frac{1}{2}\ln{|\mbf{R}^{-1}|}
}_{J_y \, \text{from} \, -\ln{p(\y|\x)}}
+ \underbrace{
  \frac{1}{2}\mbf{e}_v^T \mbf{Q}^{-1} \mbf{e}_v - \frac{1}{2}\ln{|\mbf{Q}^{-1}|}
}_{J_v \, \text{from} \, -\ln{p(\x)}},
\end{equation}
where $\mbf{e}_y \in \mathbb{R}^{MK}$ is our nonlinear measurement error, $\mathbf{R}$ is the corresponding measurement covariance matrix, and $|\cdot|$ is the matrix determinant. Similarly, $\mbf{e}_v \in \mathbb{R}^{DK}$ and $\mbf{Q}$ are the error and covariance for the prior.

Focusing on the measurements, common practice assumes the measurements are not time-correlated, making the inverse covariance (and covariance) a block-diagonal matrix $\mbf{R}^{-1} = \text{diag}(\mbf{W}_1, \mbf{W}_2, \dots, \mbf{W}_K)$. This assumption allows for the following factorization for the measurements:
\begin{equation}
J_y = \sum_{k=1}^{K} \left(\frac{1}{2} 
      \mbf{e}_{y,k}^T \mbf{W}_k \mbf{e}_{y,k} - \frac{1}{2}\ln{|\mbf{W}_k|}
    \right),
\end{equation}
where $\mbf{e}_y^T = \left[\mbf{e}_{y,1}^T \: \mbf{e}_{y,2}^T \: \cdots \: \mbf{e}_{y,K}^T \right]$. The block-diagonal sparsity for the measurements combined with the sparsity pattern of the prior results in the familiar block-tridiagonal inverse covariance of the \ac{MAP} estimate \cite{Barfoot2017}.

However, the assumption of no time correlations in the measurements is not always a valid one in real-world robotics problems. We seek to improve batch estimation performance by additionally modelling and learning the time correlations in the measurements.

\subsection{Modelling Correlations}
We model time correlations by parameterizing the inverse measurement covariance, $\mathbf{R}^{-1}$, as a symmetric block-banded matrix. The (block)-bandwidth\footnote{Throughout the rest of this paper we will drop the term ``block'' out of convenience when referring to the bandwidth.}, $b$, defines the number of non-zero blocks above and below the main block diagonal. For example, $b=0$ and $b=1$ are block-diagonal and block-tridiagonal matrices, respectively. Consequently the covariance, $\mathbf{R}$, will generally be a dense matrix, but in practice will have entries decaying to zero the further away they are from the main block diagonal \cite{demko1984decay}. By parameterizing $\mathbf{R}^{-1}$ as block-banded, rather than $\mathbf{R}$, our batch estimator can take advantage of the block-banded sparsity while correlating measurements across the entire trajectory.

Our parameterization for the inverse covariance is an upper-diagonal-lower decomposition, $\mbf{R}^{-1} = \Ss^T\Ww\Ss$, where $\mbf{W} = \text{diag}(\mbf{W}_1, \mbf{W}_2, \dots, \mbf{W}_K)$,
\begin{equation}
\Ss = \Sbig
\end{equation}
is a block-banded lower-triangular matrix with $\Ss_{k,b'} = \mbf{0}$ for $b' > b$, $k = 1, 2, \dots, K$, and $\bfone$ is the identity matrix.

This choice of parameterization factors the conditional likelihood, $p(\y|\x)$, in a convenient way. Substituting $\mbf{R}^{-1} = \Ss^T\Ww\Ss$ with bandwidth $b$ into $-\ln p(\y|\x)$,
\begin{align} \label{eq:corr_factors}
  &\frac{1}{2}\mbf{e}_y^T \mbf{R}^{-1} \mbf{e}_y - \frac{1}{2}\ln{|\mbf{R}^{-1}|} =\nonumber\\
  &\sum_{k=b+1}^K \frac{1}{2}\left( 
    \mbf{e}_{k-b:k}^T [ \Ss_{k,b:1} \; \bfone ]^T \Ww_k [ \Ss_{k,b:1} \; \bfone ] \mbf{e}_{k-b:k} 
   - \ln{|\Ww_k|} \right) \nonumber\\
  &\qquad + \frac{1}{2} \mbf{e}_{1:b}^T [ \Ss_{b,b:1} \; \bfone ]^T \Ww_b [ \Ss_{b,b:1} \; \bfone ] \mbf{e}_{1:b} 
    - \frac{1}{2}\ln{|\Ww_b|} \nonumber\\
  &\qquad + \quad \vdots \quad \text{(factors between)} \nonumber\\
  &\qquad + \frac{1}{2} \mbf{e}_{y,1}^T \Ww_1 \mbf{e}_{y,1} - \frac{1}{2}\ln{|\Ww_1|}, \\[5pt]
  & \mbf{e}_{k_1:k_2}^T = 
    \bbm \mbf{e}_{y,k_1}^T & \mbf{e}_{y,k_1+1}^T & \cdots & \mbf{e}_{y,k_2}^T \ebm, \\[5pt]
  &\Ss_{k, b':1} = \bbm \Ss_{k, b'} & \Ss_{k, b'-1} & \cdots & \Ss_{k, 1} \ebm.
\end{align}
For example with bandwidth $b=1$,
\begin{align}
  &\frac{1}{2}\mbf{e}_y^T \mbf{R}^{-1} \mbf{e}_y - \frac{1}{2}\ln{|\mbf{R}^{-1}|} =\\
  &\quad\sum_{k=2}^K \frac{1}{2}\left( 
    \mbf{e}_{k-1:k}^T [ \Ss_{k,1} \; \bfone ]^T \Ww_k [ \Ss_{k,1} \; \bfone ] \mbf{e}_{k-1:k} 
   - \ln{|\Ww_k|} \right) \nonumber\\
   &\qquad + \frac{1}{2} \mbf{e}_{y,1}^T \Ww_1 \mbf{e}_{y,1} - \frac{1}{2}\ln{|\Ww_1|}.
\end{align}
Expanding the product of the correlation parameters and the errors, $[ \Ss_{k,1} \; \bfone ] \mbf{e}_{k-1:k} = \Ss_{k,1} \mbf{e}_{y,k-1} + \mbf{e}_{y,k}$,
which we recognize to be similar to the differencing approach for handling correlations in filters \cite{bryson1968estimation}, but with the measurement errors instead of the measurements, $\y_{k-1}$ and $\y_k$.


\subsection{Learning a Constant Noise Model}
When we expect the noise characteristics to not vary throughout the trajectory, we can set $\Ww_k = \Ww_*$ and $\Ss_{k,b:1} = \Ss_*$ constants for $k = b+1, b+2, \dots, K$. 

Given a training dataset of length $N$, $\mathcal{D} = \{ \x_{1:N}, \y_{1:N}\}$\footnote{We index using $i=1,2, \dots, N$ to distinguish the training dataset from the test dataset, which uses $k = 1,2, \dots, K$.}, where $\x_{1:N}$ is the groundtruth trajectory, we can optimize for $\mbs{\theta}_*= \{ \Ww_*, \Ss_* \}$ using a maximum\footnote{We are technically minimizing the objective here since we choose to work with the negative logarithm.} likelihood objective:
\begin{align} \label{eq:constant_theta}
  \hspace{-.5em}\mathcal{L} &= 
    \hspace{-.5em}\sum_{i=b+1}^N \frac{1}{2}\left( 
    \mbf{e}_{i-b:i}^T [ \Ss_* \; \bfone ]^T \Ww_* [ \Ss_* \; \bfone ] \mbf{e}_{i-b:i} 
   - \ln{|\Ww_*|} \right), 
\end{align}
where we dropped the factors corresponding to $i=1, \dots, b$. We assume the noise models for these factors are known. In practice, they can be approximated by optimizing a similar objective to (\ref{eq:constant_theta}), but for smaller bandwidths.

We take the partial derivatives of $\mathcal{L}$ with respect to $\mbs{\theta}_*$ and set them to zero to optimize (\ref{eq:constant_theta}). Starting with $\Ss_*$,
\begin{equation}
\hspace{-.5em}\frac{\partial \mathcal{L}}{\partial \Ss_*} = 
  \hspace{-.5em}\sum_{i=b+1}^N \Ww_*\left(
     \mbf{e}_{y,i} \mbf{e}_{i-b:i-1}^T + \Ss_* \mbf{e}_{i-b:i-1} \mbf{e}_{i-b:i-1}^T
  \right). 
\end{equation}
Setting to zero, we obtain
\begin{align} \label{eq:constant_S}
 \hspace{-.5em} \Ss_* &= -\left( \sum_{i=b+1}^N \hspace{-.5em} \mbf{e}_{y,i} \mbf{e}_{i-b:i-1}^T \right) \hspace{-.5em}
    \left(\sum_{i=b+1}^N \hspace{-.5em} \mbf{e}_{i-b:i-1} \mbf{e}_{i-b:i-1}^T \right)^{-1}\hspace{-1em}. 
\end{align}
Taking the partial derivative\footnote{While $\Ww_*$ is a symmetric matrix, we do not need to take a structured derivative since we are setting the result to zero.} of $\mathcal{L}$ with respect to $\Ww_*$,
\begin{equation}
\frac{\partial \mathcal{L}}{\partial \Ww_*} = \hspace{-.5em}
  \sum_{i=b+1}^N \frac{1}{2} \left(
    [ \Ss_* \; \bfone ] \mbf{e}_{i-b:i} \mbf{e}_{i-b:i}^T [ \Ss_* \; \bfone ]^T - \Ww^{-1}
  \right).
\end{equation}
Setting to zero, we obtain
\begin{equation}
  \Ww_* = (N-b) \left(
    \sum_{i=b+1}^N [ \Ss_* \; \bfone ] \mbf{e}_{i-b:i} \mbf{e}_{i-b:i}^T [ \Ss_* \; \bfone ]^T
  \right)^{-1},
\end{equation}
where $\Ss_*$ is known from (\ref{eq:constant_S}).

\begin{figure*}[ht!]
  \vspace*{0.05in}
  \centering
  \includegraphics[trim={0.95in 4.87in 0.75in 4.75in},clip]{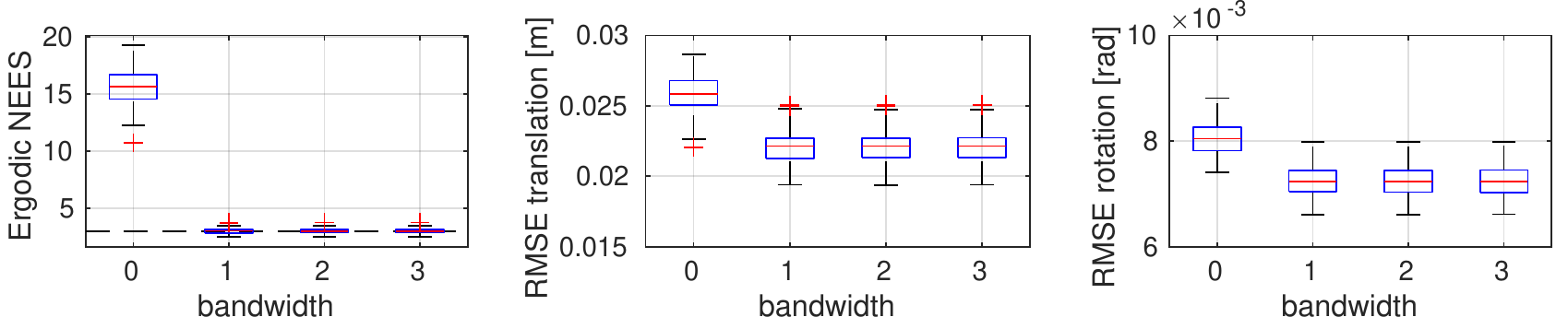}
  \caption{Simulations: Box plots of the ergodic \ac{NEES}, \ac{RMSE} translation, and \ac{RMSE} rotation over 100 simulation trials. Ideally the ergodic \ac{NEES} should be the dimension of the state, $D=3$. Not modelling the time correlations (bandwidth 0) achieves the worst performance. A bandwidth of 1 is enough to fully model the correlations in the simulated data. The proposed learning algorithm is capable of learning the correct noise properties even when the model is over-parameterized (bandwidth greater than 1 in this case).}
  \label{fig:sim}
  \vspace*{-0.21in}
\end{figure*}

\subsection{Learning a Varying Noise Model}
When the noise characteristics are expected to vary throughout the trajectory, we can formulate a prediction scheme through machine learning by introducing a feature, $\fe \in \mathbb{R}^L$, for each measurement. Given a training dataset, $\mathcal{D} = \{ \x_{1:N}, \y_{1:N}, \fe_{1:N}\}$, our noise model for a target feature, $\fe_*$, at test time is
\begin{equation} \label{eq:predictor}
  \mbs{\theta}_* = \mbs{\theta}(\fe_*|\mathcal{D}) = \{ \Ww(\fe_*|\mathcal{D}), \Ss(\fe_*|\mathcal{D}) \}.
\end{equation}
We choose to apply the local kernel estimation approach of Vega-Brown et al. \cite{vega2014nonparametric} and adapt it to predict our correlated noise parameters. We first review their methodology for uncorrelated noise prediction (i.e., $\mbs{\theta}(\fe_*|\mathcal{D}) = \Ww(\fe_*|\mathcal{D})$), then present how we adapt their methodology.

Vega-Brown et al. \cite{vega2014nonparametric} formulate an estimator for $\mbs{\theta}_*$ given a target feature, $\fe_*$. A joint posterior between $\mbs{\theta}_*$ and $\mbs{\theta}_{1:N}$ is formed using Bayes' rule and $\mbs{\theta}_{1:N}$ is then marginalized out, resulting in
\begin{equation} \label{eq:cello_posterior}
  \hspace{-.5em}p(\mbs{\theta}_*| \fe_*, \mathcal{D}) \propto 
  \left(
    \prod_{i=1}^N p(\y_i|\x_i, \mbs{\theta}_*, \fe_i, \fe_*)
  \right) p(\mbs{\theta}_*|\fe_*),
\end{equation}
where we note the uncorrelated factorization of the likelihood terms. They show that an Inverse-Wishart distribution can be applied as the prior on the covariance, which results in a Student's t distribution over the noise \cite{peretroukhin2016probe}. In our work we will apply an uninformative prior, $p(\mbs{\theta}_*|\fe_*) \propto 1$, which is an adaptation of their prior work, CELLO \cite{vega2013cello,vega2013celloem}.

The key innovation of Vega-Brown et al. \cite{vega2014nonparametric} is their choice in modelling the likelihood in (\ref{eq:cello_posterior}) (referred to as the extended likelihood), as
\begin{equation}
p(\y_i|\x_i, \mbs{\theta}_*, \fe_i, \fe_*) \propto p(\y_i|\x_i, \mbs{\theta}_*)^{\ke(\fe_i, \fe_*)},
\end{equation}
where $\ke(\fe_i, \fe_*)$ is a kernel function. The extended likelihood is modelled using the known likelihood, $p(\y_i|\x_i, \mbs{\theta}_*)$. Motivated by enforcing a notion of smoothness, they derived this model by bounding the information divergence between the two likelihood distributions. We refer to Vega-Brown et al. \cite{vega2014nonparametric} for a more detailed explanation.

We adapt their methodology for time-correlated noise by applying the correlated factorization scheme shown in (\ref{eq:corr_factors}), replacing the uncorrelated factorization in (\ref{eq:cello_posterior}):
\begin{align} \label{eq:corr_posterior}
  p(&\mbs{\theta}_*| \fe_*, \mathcal{D}) \propto 
    p(\y_{1:b}|\x_{1:b}, \fe_{1:b}) \\
    & \times \left(\prod_{i=b+1}^{N}
      p(\y_{i}|\y_{i-b:i-1}, \x_{i-b:i}, \mbs{\theta}_{*}, \fe_{i}, \fe_{*})
    \right)p(\mbs{\theta}_*|\fe_*). \nonumber
\end{align}
We use the negative logarithm of (\ref{eq:corr_posterior}) as our training objective, optimizing for $\mbs{\theta}_* = \{\Ww_*, \Ss_* \}$. As mentioned, we will apply an uninformative prior, $p(\mbs{\theta}_*|\fe_*) \propto 1$. Modelling the extended likelihood using our known likelihood and kernel function, the objective function is
\begin{align} \label{eq:vary_theta}
\hspace{-.7em} \mathcal{L}_{\ke} =
  \hspace{-.7em} \sum_{i=b+1}^N \hspace{-.2em} \frac{\ke_i}{2} \hspace{-.2em} \left(
    \mbf{e}_{i-b:i}^T [ \Ss_{*} \; \bfone ]^T \Ww_* [ \Ss_{*} \; \bfone ] \mbf{e}_{i-b:i} 
   - \ln{|\Ww_*|} \right), \hspace{-.4em}
\end{align}
where $\ke_i = \ke(\fe_i, \fe_*)$ and, similar to (\ref{eq:constant_theta}), we drop the lower bandwidth factors corresponding to $i = 1, 2, \dots, b$. The resulting objective (\ref{eq:vary_theta}) is similar to (\ref{eq:constant_theta}), but now with each term weighted by evaluations of the kernel, $\ke_i$. Taking the partial derivatives and setting them to zero,
\begin{align} \label{eq:vary_S}
\Ss(\fe_*|\mathcal{D}) &=  
  -\left( \sum_{i=b+1}^N  \ke_i\mbf{e}_{y,i} \mbf{e}_{i-b:i-1}^T \right) \nonumber\\ 
    &\qquad\times  \left(\sum_{i=b+1}^N  \ke_i\mbf{e}_{i-b:i-1} \mbf{e}_{i-b:i-1}^T \right)^{-1}, \\
\hspace{-.5em} \Ww(\fe_*|\mathcal{D}) \hspace{-.1em} &= \hspace{-.1em} \left( \frac{1}{H} \hspace{-.2em}
    \sum_{i=b+1}^N \hspace{-.2em} \ke_i[ \Ss_* \; \bfone ] \mbf{e}_{i-b:i} \mbf{e}_{i-b:i}^T [ \Ss_* \; \bfone ]^T \hspace{-.1em}
  \right)^{-1} \hspace{-1.1em}, \hspace{-.1em}\label{eq:vary_W}
\end{align}
where $H = \sum_{i=b+1}^N \ke(\fe_i, \fe_*)$. For our work we use the squared exponential kernel,
\begin{equation}
  \ke(\fe_i, \fe_*) = \exp{\left(
    -\frac{1}{2} (\fe_i - \fe_*)^T \mbf{M} (\fe_i - \fe_*)
  \right)}.
\end{equation}
The weight matrix, $\mathbf{M}$, is a hyperparameter that we train via maximum likelihood as suggested by Vega-Brown et al. \cite{vega2013cello}.


\begin{figure}[h]
  \centering
  \includegraphics[width=0.45\textwidth, trim={0in 1.2in 0in 0in}, clip]{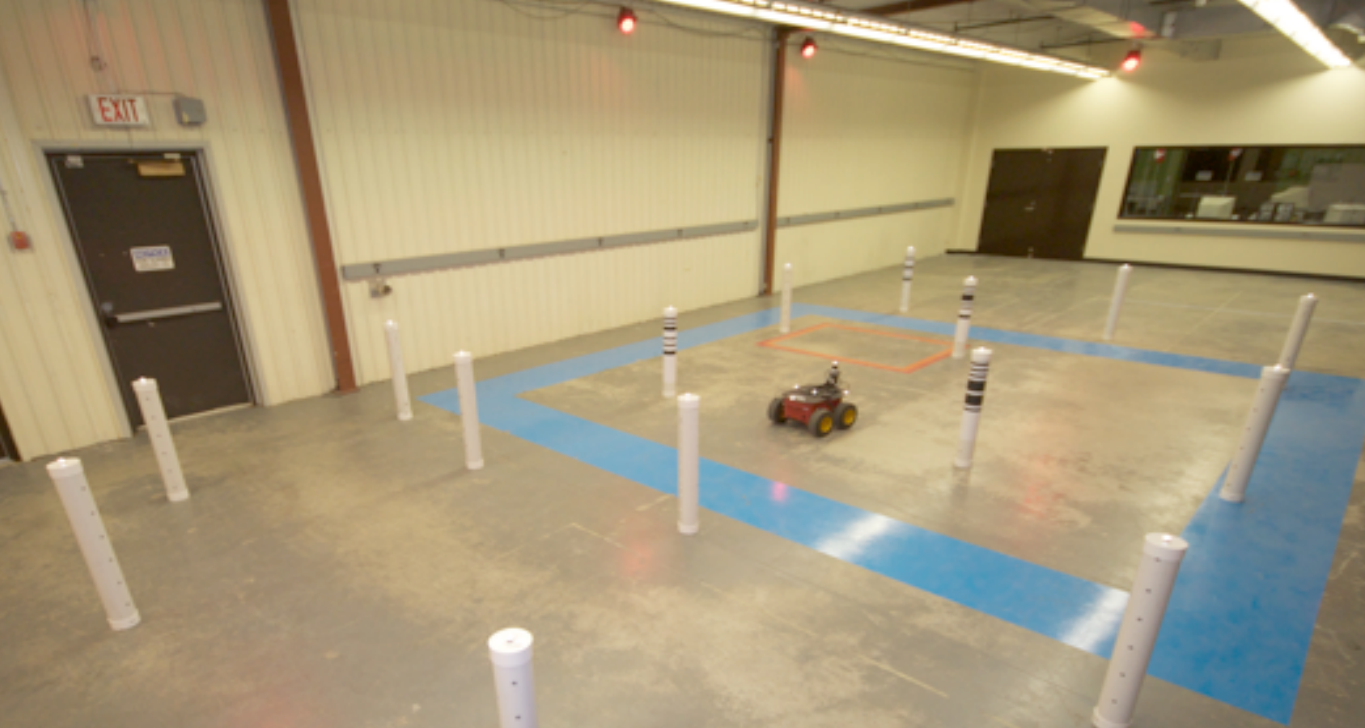}
  \caption{The dataset for demonstrating the proposed method was acquired by a robot with a laser rangefinder and wheel odometers that was driven around tubular landmarks. A Vicon motion capture system provides the groundtruth.}
  \label{fig:experiment}
  \vspace*{-0.22in}
\end{figure}

\begin{table*}[ht]
  \vspace*{0.05in}
  \centering
  \caption{Experiments: Quantitative results on the dataset with the best results in bold font. Note that the ideal value for the ergodic \ac{NEES} in this problem is 3, the dimension of the state. The trend is that the SVD-FEAT-X methods perform the best, with larger bandwidth settings generally performing better.}
  \label{tab:exp}
  \begin{tabular}{ l ? c c c c c ? A A A A c ? A A A A c }
    \toprule
    \midrule
      & \multicolumn{5}{c|}{Ergodic NEES} 
      & \multicolumn{5}{c|}{RMSE translation [m]} 
      & \multicolumn{5}{c}{RMSE rotation [rad]}\\
    \midrule
      \textbf{Folds} & 01 & 02 & 03 & 04 & {\textbf{AVG}} 
        & 01 & 02 & 03 & 04 & {\textbf{AVG}} 
        & 01 & 02 & 03 & 04 & {\textbf{AVG}} \\
    \midrule
      P2P-CONST & 15.4 & 15.4 & 19.5 & 15.9 & \color{blue}{16.6}
                  & 0.0215 & 0.0211 & 0.0240 & 0.0226 & \color{blue}{0.0223}
                  & 0.0175 & 0.0167 & 0.0188 & 0.0149 & \color{blue}{0.0169} \\
      SVD-CONST  & 12.4  & 16.2 & 16.9  & 12.9 & \color{blue}{14.6} 
                 & 0.0219 & 0.0225 & 0.0241 & 0.0226 & \color{blue}{0.0228}
                 & 0.0185 & 0.0187 & 0.0198 & 0.0179 & \color{blue}{0.0187} \\
      SVD-FEAT-0  & 11.4 & 14.1 & 15.0 & 13.5 & \color{blue}{13.5} 
                 & 0.0213 & 0.0215 & 0.0233 & 0.0224 & \color{blue}{0.0221}
                 & 0.0150 & 0.0144 & \textbf{0.0164} & 0.0150 & \color{blue}{0.0152} \\
      SVD-FEAT-1  & 4.53 & 5.25 & 5.64 & 5.85 & \color{blue}{5.32}
                 & \textbf{0.0196} & 0.0196 & 0.0218 & 0.0218 & \color{blue}{0.0207}
                 & 0.0134 & \textbf{0.0128} & \textbf{0.0164} & 0.0121 & \color{blue}{0.0137} \\
      SVD-FEAT-3  & 3.71 & 3.85 & 4.77 & 4.61 & \color{blue}{4.23}
                 & \textbf{0.0196} & 0.0193 & 0.0219 & \textbf{0.0217} & \color{blue}{0.0206}
                 & \textbf{0.0131} & 0.0130 & 0.0165 & 0.0119 & \color{blue}{\textbf{0.0136}} \\
      SVD-FEAT-5  & \textbf{3.41} & \textbf{3.37} & \textbf{4.44} & \textbf{4.42} & \color{blue}{\textbf{3.91}}
                 & \textbf{0.0196} & \textbf{0.0189} & \textbf{0.0216} & 0.0218 & \color{blue}{\textbf{0.0205}}
                 & 0.0133 & 0.0129 & 0.0166 & \textbf{0.0117} & \color{blue}{\textbf{0.0136}} \\
    \midrule
    \bottomrule

  \end{tabular}
  \vspace*{-0.05in}
\end{table*}

\begin{figure*}[ht]
  \centering
  \includegraphics[trim={0.95in 4.82in 0.75in 4.85in},clip]{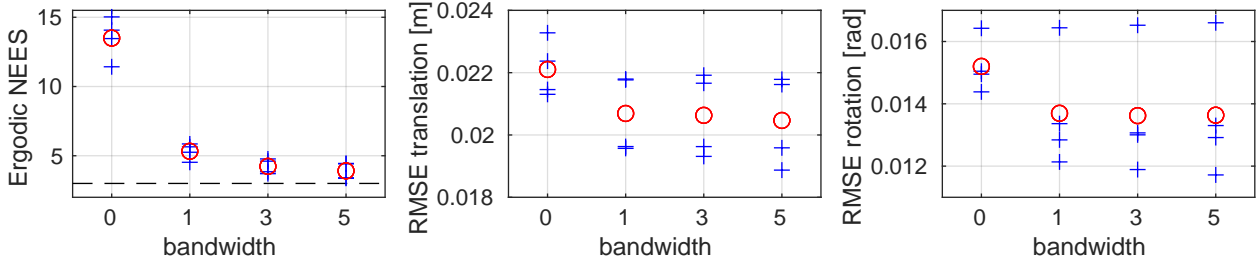}
  \caption{Experiments: Plots of the results in Table~\ref{tab:exp} for SVD-FEAT-X methods. A `+' marks an evaluation of a single fold, and a `o' indicates the average. The biggest improvements are gained by changing the bandwidth to $b=1$ from $b=0$ (i.e., no time correlations). In general we still see minor improvements for $b>1$, the exception being the rotation error for fold $03$. However, the average rotation error still appears to decrease and stabilize for larger $b$.}
  \label{fig:real}
  \vspace*{-0.2in}
\end{figure*}

\section{Experimental Results}

\subsection{Dataset}
A robot with a 2D laser rangefinder and wheel odometers is driven around tubular landmarks (see Figure~\ref{fig:experiment}). The groundtruth for the robot trajectory and landmark positions are accurately provided by a Vicon motion capture system. The rangefinder measurements are preprocessed into 2D point landmark measurements. The dataset is approximately 12000 timesteps in length with a discrete-time sampling period of $\triangle t = 0.1 s$. This dataset has been used before for demonstrating advances in batch estimation \cite{barfoot_rss14,barfoot_ijrr20}.

We can use the groundtruth to check for time correlations in the measurements. We stack the landmark measurements over a short window of 8 timesteps without overlap and calculate the sample measurement covariance\footnote{One may be tempted to use this covariance for estimation by grouping measurements, but correlations will be neglected (grouping without overlap) or measurements will be double-counted (grouping with overlap).}. The magnitude of the resulting covariance and its inverse are visualized in Figure~\ref{fig:lm_sample_cov}, clearly showing signs of time correlations (i.e., they do not appear to be block-diagonal). A similar experiment for the velocity measurements reveals that they are also time-correlated, but not as severely as the landmark measurements. 

\subsection{Problem Setup}
We perform localization as a batch state estimation problem, where the data association and true landmark locations (i.e., the map) are assumed to be known. Our state is $\x = \{\T_1, \T_2, \dots, \T_K \}$, where $\T_k = \T_{k,i} \in SE(2)$ is the relative transform between the robot frame at time $k$ and a fixed reference frame, $i$. We use the odometer measurements in a constant-velocity error function (motion model),
\begin{equation} \label{eq:motion}
\mbf{e}_{v,k} = \ln{\left( 
    \exp{ \left( \triangle t \mbf{v}_k^\wedge \right)} \T_{k-1} \T_k^{-1}
  \right)}^\vee,
\end{equation}
where $\mbf{v}_k^T = [ -v_k \quad 0 \quad -\omega_k ]$ are the forward, lateral, and rotational speeds\footnote{The negative sign is from our convention for the exponential map \cite{Barfoot2017}.} in the robot frame, and $\exp{(\cdot)}$ and $\ln{(\cdot)}$ are the $SE(2)$ exponential map and its inverse, respectively.

There are a varying number of landmarks measured at each timestep throughout the dataset. Learning a correlated noise model at the landmark-measurement level will require tedious bookkeeping of the observed landmarks and ignore potential correlations between landmarks at a single timestep. We instead preprocess the measurements into pose pseudomeasurements, $\T_{mk} = \T_{mk,i} \in SE(2)$, and learn a varying noise model for the pseudomeasurements. This is a common approach for learning noise models for rich sensor data \cite{landry2019cello,liu2018deep,russell2021multivariate,torroba2020pointnetkl,de2022deep,wong_ral20b}. We compute the pseudomeasurements using \ac{SVD} and use
\begin{equation}
  \mbf{e}_{y,k} = \ln{\left( 
     \T_{mk} \T_k^{-1}
  \right)}^\vee.
\end{equation}
We learn our noise model by defining a feature, $\fe_k \in \mathbb{R}^6$, that comprises the number of visible landmarks and their spread\footnote{2D sample mean and covariance of the landmark point measurements.} in the robot frame at time $k$.

\begin{figure*}[ht]
  \vspace*{0.05in}

  \includegraphics[trim={2.2in 4.75in 2.9in 4.85in},clip]{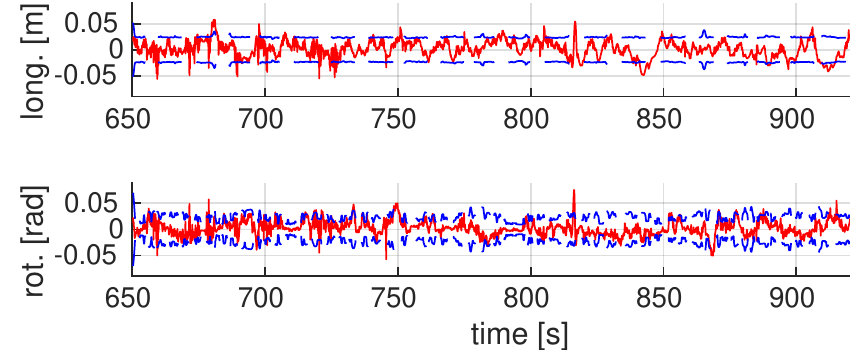}
  \includegraphics[trim={2.2in 4.75in 2.9in 4.85in},clip]{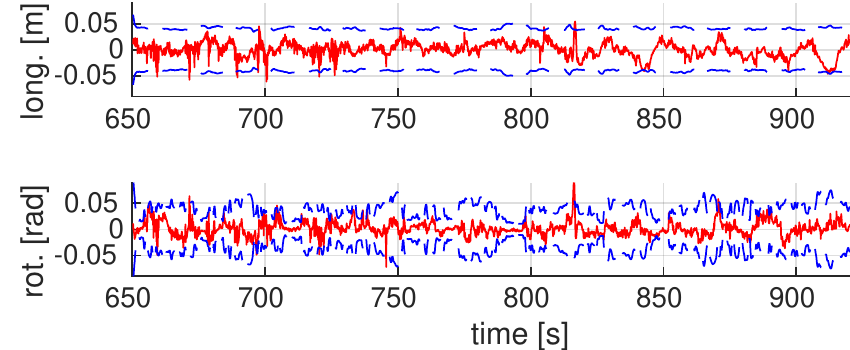}

  \caption{Experiments: Shown above are error plots with a $3\sigma$ uncertainty envelope. We plot the longitudinal translation errors and the rotational errors. We omit the lateral translation errors in the interest of space. SVD-FEAT-0 (left column) does not model any time correlations. SVD-FEAT-5 (right column), which learns an inverse measurement covariance with a block-bandwidth of 5, appears to have its errors more contained within the uncertainty envelope.}
  \label{fig:error_plot}
  \vspace*{-0.2in}
\end{figure*}

\subsection{Simulation}
We simulate multiple trials using a Bayesian experiment setup \cite{Barfoot2017} and evaluate consistency using the \ac{NEES} $\chi^2$ statistical test \cite{barshalom,chen2018weak}. Measurements are simulated by
\begin{equation}
  \y_k = \mbf{g}(\x_k, \mbf{\ell}) + \mbf{w}_k, \quad \mbf{w}_k = \Ss' \mbf{w}_{k-1} + \mbf{n},
\end{equation}
where $\y_k$ is a stacked vector of all landmark measurements, $\mbf{g}(\cdot, \mbf{\ell})$ is a nonlinear model that computes landmark measurements in the robot frame at pose $\x_k$, $\mbf{\ell}$ are the true landmark positions (i.e., map), and $\mbf{n} \sim \mathcal{N}(\mbf{0}, \mbf{R}')$. Measurements outside a $270^\circ$ field-of-view and greater than a maximum range of $5 m$ are not counted, varying the number of visible landmarks at each timestep\footnote{These specifications are similar to the real dataset.}. The noise parameters are set as $\mbf{R}' = (0.03 m)^2 \times \bfone$ and $\mbf{S}' = 0.9 \times \bfone$. For each simulation trial we generate a training sequence for the noise model and a test sequence for evaluation.

The marginal \ac{NEES} at time $k$ is computed as $\epsilon_{x,k} = \mbf{e}_{x,k}^T \hat{\mbf{P}}_k^{-1} \mbf{e}_{x,k}$,
where $\mbf{e}_{x,k}$ is the marginal error and $\hat{\mbf{P}}_k$ is the marginal posterior covariance at time $k$. The \ac{NEES}  $\chi^2$ statistical test aggregates $\epsilon_{x,k}$ over all $N_t$ trials and checks if the result is within a lower and upper bound:
\begin{equation}
  \mathcal{Q}_{\chi^2(N_t D)}(\ell) \leq \sum_{j=1}^{N_t} \epsilon_{x,k}^j \leq \mathcal{Q}_{\chi^2(N_tD)}(u),
\end{equation}
where $\ell$ and $u$ are lower and upper confidence bounds, respectively, and $\mathcal{Q}_{\chi^2(N_t D)}(\cdot)$ is the quantile function for the $\chi^2$ distribution with $N_t D$ degrees of freedom ($D=3$ is the dimension of the state).

Simulating $N_t = 100$ trials, we apply our proposed approach for the \ac{SVD} pseudomeasurements. With bandwidth $b=1$, $0.2022 \%$ of 3000 test timesteps fall outside a $99.8 \%$ confidence interval ($\ell=0.001$, $u=0.999$), which is reasonable. Using a $95 \%$ confidence interval ($\ell=0.025$, $u=0.975$), $5.09 \%$ fall outside the bounds, which is also reasonable. Similar results are produced for $b > 1$, demonstrating that we can over-parameterize and still learn an accurate covariance model\footnote{Our experiments show that more training data is required for over-parameterized settings of the bandwidth. In other words, performance may be degraded if $b$ is set too large with insufficient training data.}. When $b=0$ (i.e., original CELLO \cite{vega2013cello}), none of the trials fall within reasonable confidence intervals because the resulting estimator is overconfident.

\begin{figure}[ht]
  \centering
  \includegraphics[trim={2.6in 4.87in 2.9in 4.87in},clip]{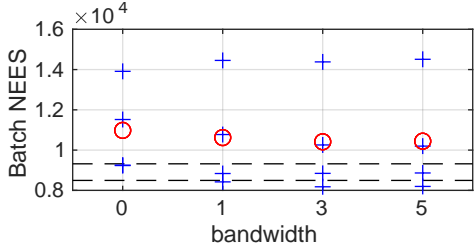}

  \caption{Experiments: The batch \ac{NEES} for SVD-FEAT-X with varying bandwidths. A `+' marks a single fold, and a `o' marks the average. Dashed lines indicate the $99.8 \%$ confidence interval, which most points do not meet. Meeting this confidence interval is the subject of future work.}
  \label{fig:batch_nees}
  \vspace*{-0.2in}
\end{figure}

We can also take an \textit{ergodic} assumption and compute the average \ac{NEES} over the entire test sequence in each trial: $\frac{1}{K} \sum_{k=1}^{K} \epsilon_{x,k}$.
The ergodic \ac{NEES} does not take into account the time correlations in the estimated posterior and thus is not valid for a statistical test. It is still useful to compute as the value on average should match the dimension of the state, $D=3$. We plot the ergodic \ac{NEES} as a box plot in Figure~\ref{fig:sim} for varying bandwidths, $b$, along with box plots of the \ac{RMSE} for translation and rotation. Performance on all three metrics is poor when $b=0$. When $b = 1, 2,$ and $3$, performance is improved and similar. 

\subsection{Real Data}
We divide the dataset into 4 equally-sized folds ($\sim3000$ timesteps). We test on each fold by training on the 3 other folds. We compare the following methods:
\subsubsection{P2P-CONST} Replaces the SVD pseudomeasurements with the original 2D landmark measurements.
\subsubsection{SVD-CONST} Learn a constant covariance for the SVD pseudomeasurements (no correlations are modelled).
\subsubsection{SVD-FEAT-X} Learn varying noise parameters for the SVD pseudomeasurements with bandwidth $b=X$.

For P2P-CONST and SVD-CONST, we train a constant covariance for the motion model in (\ref{eq:motion}). For SVD-FEAT-X, we also band the motion model covariance by $b=X$ to model time correlations in the velocity.

Table~\ref{tab:exp} shows our quantitative results. The best performers are the SVD-FEAT-X methods, with larger bandwidth settings fairing better in general. Figure~\ref{fig:real} plots the results of the SVD-FEAT-X methods, visualizing a general trend in performance improvement. Figure~\ref{fig:error_plot} qualitatively compares error plots between SVD-FEAT-0 and SVD-FEAT-5. The errors appear to be more contained within the $3\sigma$ uncertainty envelope when the time correlations are modelled.

\begin{table}[ht]
  \centering
  \caption{Timing results on SVD-FEAT-X for varying bandwidth.}.
  \label{tab:run}
  \begin{tabular}{ l ? c c c c c c }
    \toprule
    \midrule
     Bandwidth  & 0 & 1 & 2 & 3 & 4 & 5 \\
    \midrule
    Predict & 2.35s & 3.05s & 3.66s & 4.06s & 4.51s & 4.74s \\
    Optimize & 0.85s & 1.00s & 1.07s & 1.34s & 1.47s & 1.62s \\
    \midrule
    \bottomrule
  \end{tabular}
  \vspace*{-0.2in}  
\end{table}

\section{Discussion and Future Work}
We present improvements in batch estimation for both the posterior mean and covariance. We demonstrate consistency using a statistical test in a controlled simulation and show improvements to the ergodic \ac{NEES} on real data, both of which evaluate at the marginal level. For future work, we will focus on achieving consistency using real data through a statistical test. In theory, computing the \textit{batch} \ac{NEES} over the entire trajectory should account for the correlations in the posterior, in contrast to the ergodic \ac{NEES}, and be valid for a $\chi^2$ statistical test. The batch \ac{NEES} is $\epsilon = \mbf{e}^T \hat{\mbf{P}}^{-1} \mbf{e}$,
where $\hat{\mbf{P}}$ is the posterior covariance of the entire trajectory and $\mbf{e}$ is the full-trajectory error. Figure~\ref{fig:batch_nees} shows the batch \ac{NEES} using the SVD-FEAT-X methods on our dataset with the $99.8 \%$ confidence interval indicated by the dashed lines. Our methods tend to still be overconfident on some folds, suggesting there are remaining nonidealities for which we must account. One such nonideality might be that the groundtruth (i.e., the Vicon motion capture) is not truly perfect. We may need to account for this uncertainty in both training and evaluation. Unsupervised training using \ac{EM} \cite{shumway1982approach,vega2013celloem} may be of value here and is also an item for future work.

The average wall-clock times for the SVD-FEAT-X methods are shown in Table~\ref{tab:run} for a 3000 timestep test sequence. `Predict' evaluates the noise parameters for every timestep (see (\ref{eq:vary_S}), (\ref{eq:vary_W})). `Optimize' runs \ac{GN} until convergence and Takahashi's method \cite{takahashi73} for efficiently computing the marginal posterior covariances. The wall-clock times appear to increase linearly by bandwidth. However, even the slowest times are reasonable given the sequence length. There is potential for online application via a sliding window, which we leave as future work targeting larger-scale trajectories and richer sensor data.

\section{Conclusion}
We propose a method for modelling and learning a time-correlated noise model. Our method is applicable to batch state estimation and is capable of modelling correlations over long time periods. We train our models using the groundtruth trajectory and show improvements in estimator performance in simulation and on an experimental dataset.






\section*{ACKNOWLEDGMENT}
We thank the Natural Sciences and Engineering Research Council of Canada (NSERC) for supporting this work.


\newpage
\bibliographystyle{bib/IEEEtran}
\bibliography{bib/refs}

\end{document}